\documentclass[letterpaper]{article} % DO NOT CHANGE THIS
\usepackage{aaai2026}  % DO NOT CHANGE THIS
\usepackage{times}  % DO NOT CHANGE THIS
\usepackage{helvet}  % DO NOT CHANGE THIS
\usepackage{courier}  % DO NOT CHANGE THIS
\usepackage[hyphens]{url}  % DO NOT CHANGE THIS
\usepackage{graphicx} % DO NOT CHANGE THIS
\usepackage{natbib}  % DO NOT CHANGE THIS AND DO NOT ADD ANY OPTIONS TO IT
\usepackage{caption} % DO NOT CHANGE THIS AND DO NOT ADD ANY OPTIONS TO IT
\usepackage{algorithm}
\usepackage{algorithmic}

\usepackage{newfloat}
\usepackage{listings}
\DeclareCaptionStyle{ruled}{labelfont=normalfont,labelsep=colon,strut=off} % DO NOT CHANGE THIS
\floatstyle{ruled}
\newfloat{listing}{tb}{lst}{}
\floatname{listing}{Listing}
\usepackage{booktabs}
\usepackage{amsmath}
\usepackage[table]{xcolor}
\usepackage{makecell}
\usepackage{amssymb}
\usepackage{bm}  
\title{Adaptive LiDAR Scanning: Harnessing Temporal Cues for Efficient 3D Object Detection via Multi-Modal Fusion}
\author{
    Sara Shoouri, Morteza Tavakoli Taba, Hun-Seok Kim
}
\affiliations{
    University of Michigan\\
    \{sshoouri, tmorteza, hunseok\}@umich.edu
}

\usepackage{bibentry}
\begin{document}

\maketitle

\begin{abstract}
Multi-sensor fusion using LiDAR and RGB cameras significantly enhances 3D object detection task. However, conventional LiDAR sensors perform dense, stateless scans, ignoring the strong temporal continuity in real-world scenes. This leads to substantial sensing redundancy and excessive power consumption, limiting their practicality on resource-constrained platforms. To address this inefficiency, we propose a predictive, history-aware adaptive scanning framework that anticipates informative regions of interest (ROI) based on past observations. Our approach introduces a lightweight predictor network that distills historical spatial and temporal contexts into refined query embeddings. These embeddings guide a differentiable Mask Generator network, which leverages Gumbel-Softmax sampling to produce binary masks identifying critical ROIs for the upcoming frame. Our method significantly reduces unnecessary data acquisition by concentrating dense LiDAR scanning only within these ROIs and sparsely sampling elsewhere. %For end-to-end optimization, we introduce two key innovations: (1) a differentiable voxelization approach that allows gradient propagation through traditionally non-differentiable voxel quantization and (2) a risk-averse Conditional Value-at-Risk (CVaR) loss to ensure robust performance on critical, small-scale objects. 
Experiments on nuScenes and Lyft benchmarks demonstrate that our adaptive scanning strategy reduces LiDAR energy consumption by over $65\%$ while maintaining competitive or even superior 3D object detection performance compared to traditional LiDAR-camera fusion methods with dense LiDAR scanning. %Our codes and models will be publicly available upon acceptance.
\end{abstract}
%\vspace{-5mm}
 
\vspace{-4mm}
\section{Introduction}
\label{sec:intro}
Multisensor fusion for 3D object detection leverages the complementary strengths of cameras and LiDAR to improve perception reliability in autonomous driving \cite{yan2023cross,bai2022transfusion, li2022unifying, liu2023bevfusion, chen2023futr3d}. Cameras capture rich semantic cues, such as color and texture, while LiDAR provides spatial information through direct depth measurements. By fusing these modalities, autonomous systems significantly improve their reliability to handle challenging scenarios \cite{bai2022transfusion}.

Despite the sophistication achieved in modern fusion techniques, fundamental inefficiency persists at the point of LiDAR data acquisition. Conventional LiDAR sensors operate in a `stateless' or `memoryless' manner, performing dense uniform-angle scans at each frame as if observing the scene for the very first time.  This approach ignores the strong temporal continuity inherent in real-world environments, where the world does not completely rearrange itself every tenth of a second. For instance, studies in data compression demonstrate that the static background is highly predictable from prior frames. Consequently, only a small fraction of the points corresponding to dynamic objects require explicit updates \cite{feng2020real}. By repeatedly re‐scanning unchanged areas with full resolution, LiDAR wastes a significant portion of its energy on low‐value measurements.
\begin{figure}[t]
    \centering
    \includegraphics[width=0.75\columnwidth]{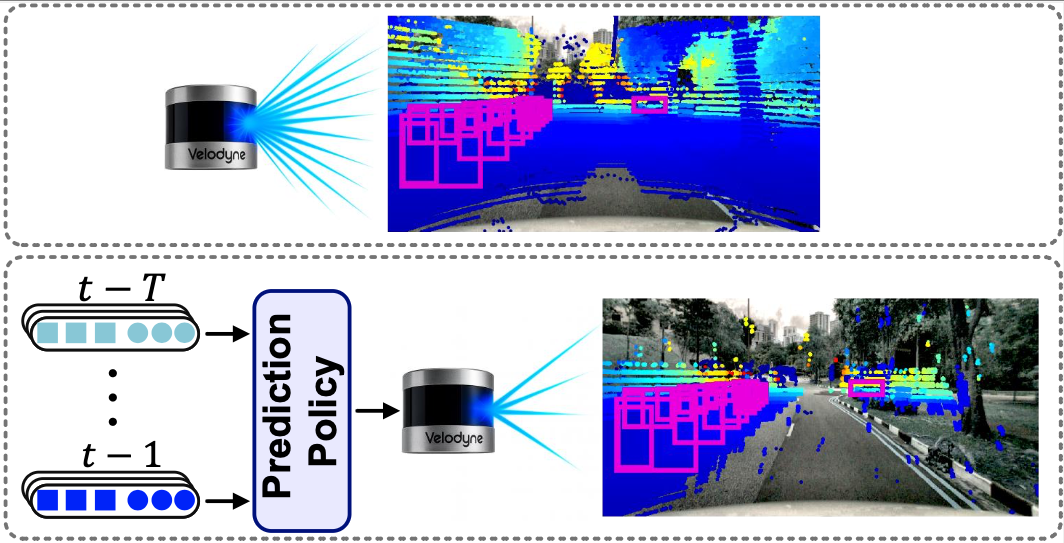}
    \vspace{-3mm}
    \caption{Top: Conventional uniform LiDAR scanning. Bottom: Our adaptive LiDAR scanning, which leverages past frames to predict the next frame’s ROIs. ROIs are scanned densely and non-ROI areas receive sparse sampling.}
    \vspace{-7mm}
    \label{fig:fig_intro}
\end{figure}
Such redundancy imposes a substantial burden on the sensing system’s power budget, a critical constraint for small form factor integration.
This power demand is especially evident when comparing LiDAR with other passive sensors: a typical automotive camera requires only $1–5 W$ per unit for capture and preprocessing \cite{sahin2019long,ambarella_cv22aq}, whereas a LiDAR, as an active sensor generating beam-steered laser pulses, requires significantly more power in the range of $10 - 100 W$ \cite{velodyne_hdl32e,velodyne_hdl64e}. %LiDAR's energy consumption arises from several key functions: generating laser pulses, steering the beam, and processing the return signals \cite{lee2020accuracy, raj2020survey, li2022progress, tayebati2025generative}. 
Laser emission is particularly power-demanding as each pulse must be strong enough to reach a target and return with sufficient intensity to be detected. Extending detection range forces pulse energy to rise steeply as return strength decays roughly with the fourth power of distance. Moreover, achieving finer angular resolution requires larger optics or higher launch power \cite{lee2020accuracy, raj2020survey, tayebati2025generative}. In real‐world deployments, these factors translate into substantial per‐scan energy demands. The Velodyne HDL-32E LiDAR used for the nuScenes dataset \cite{caesar2020nuscenes} consumes roughly $12 W$, translating to approximately $0.6\,\mathrm{J}$ per full scan at 20 Hz \cite{velodyne_hdl32e}, whereas the Velodyne HDL-64E used in Lyft dataset \cite{lyft2019, houston2021one} draws about $60 W$, or approximately $6\,\mathrm{J}$ per full scan at $10$ Hz \cite{velodyne_hdl64e}. Consequently, sustaining such high power at typical frame rates is prohibitive for power-constrained sensing platforms.

These limitations demand a paradigm shift from fixed-resolution scanning to a predictive, history-aware adaptive sensing strategy. We propose that an intelligent system can anticipate regions of interest (ROI) by leveraging the memory from recent past observations. Rather than treating each frame in isolation, our method leverages a sequence of learned historical query embeddings, which encode recent spatial and temporal context, to forecast where the most salient information (or ROIs) will be in the subsequent frame. This allows the system to proactively allocate its resources for intelligent adaptive data acquisition.

To implement this concept, we introduce a two-stage adaptive scanning LiDAR and camera fusion pipeline. First, a lightweight predictor ingests historical object queries and produces refined embeddings $Q'$ that anticipate the spatial distribution of dynamic actors in the upcoming frame. Second, these predictions are fed into a differentiable Mask Generator, which produces a binary mask via the Gumbel-Softmax trick \cite{jang2016categorical} over the LiDAR's field of view. This mask defines the critical ROIs. The final scan pattern is dictated by this mask, activating dense LiDAR scanning within ROIs only while scanning sparsely elsewhere. By complementing this sparse LiDAR sampling with surround-view RGB camera images, we preserve detection performance even in areas with sparse LiDAR scans.

We introduce two key techniques to enable end-to-end optimization of this pipeline. First, a differentiable voxelization method overcomes the limitations of traditional voxelization, which is inherently non-differentiable and blocks gradient flow. Our approach provides heuristic gradients that map voxel-level losses back to the point-level inputs, enabling end-to-end training of the entire system with adaptive non-uniform scanning. Second, we further employ a Conditional Value-at-Risk (CVaR) loss \cite{rockafellar2000optimization} to train the scanning Mask Generator. This risk-averse objective forces the model to prioritize accurate mask generation even for small yet critical objects like pedestrians, ensuring the system remains robust in high-risk scenarios. Experimental results on the nuScenes and Lyft benchmarks confirm the effectiveness of our framework. %It achieves over a $65\%$ reduction in LiDAR power consumption without sacrificing the accuracy of 3D object detection and, in some cases, even improving the performance.

Our contributions can be summarized as follows:
\vspace{-2mm}

\begin{itemize}
  \item We introduce a history-driven adaptive LiDAR paradigm that departs from traditional memoryless acquisition, using past observations to forecast future ROI positions for dynamic sensing resource allocation.
  \item We propose a transformer-based LiDAR-camera fusion model integrating a history-aware Query Predictor module and a differentiable Mask Generator network trained with a risk-averse CVaR, effectively converting predictions into adaptive spatial scanning patterns.
 \item We facilitate a custom differentiable voxelization layer to enable end-to-end training of our adaptive sensing fusion model. Empirical evaluation on the large-scale nuScenes and Lyft datasets achieves over $65\%$ reduction in LiDAR scanning density, while consistently matching or surpassing conventional dense-scan 3D detection methods.
\end{itemize}
\begin{figure*}[!t]
 \centering
\includegraphics[width=2.0\columnwidth]{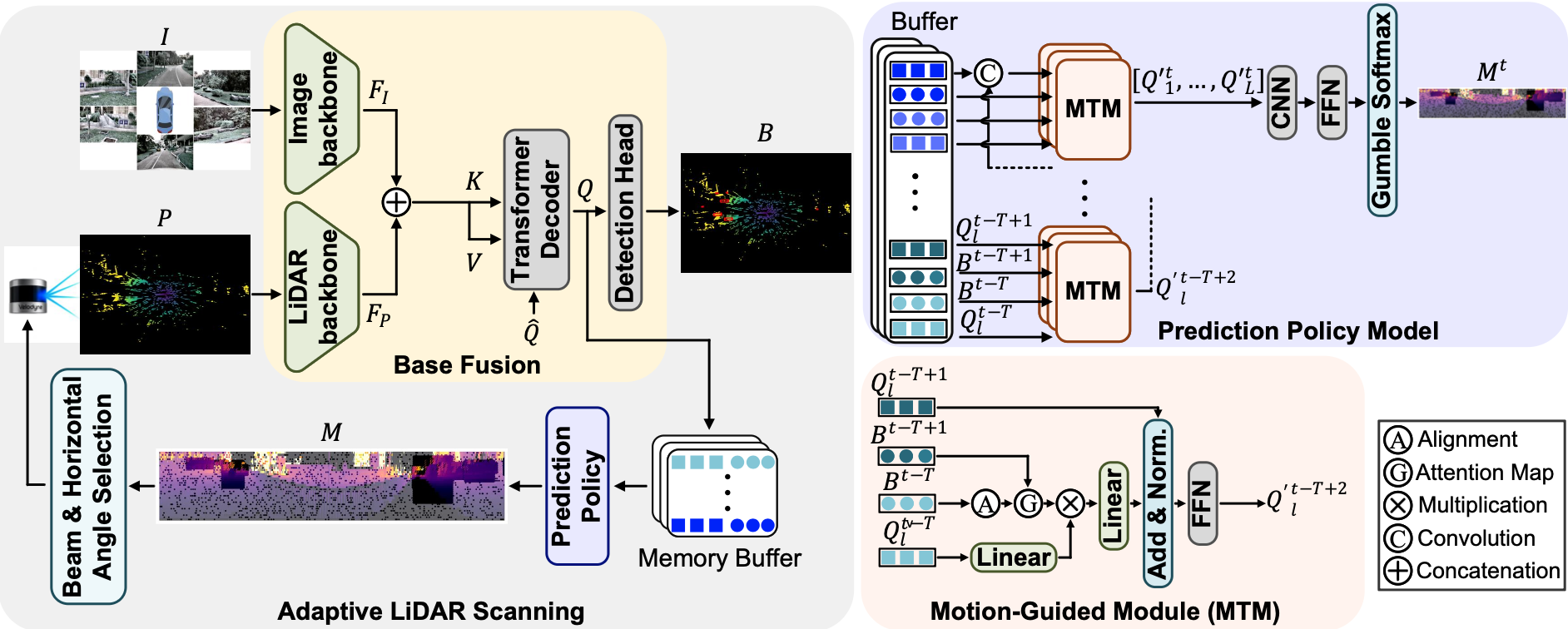}
 %\vspace{-0.75\baselineskip}
 \centering
 \vspace{-2mm}
 \caption{(a) Overview of our adaptive LiDAR scanning pipeline. Prediction model utilizes historical information to identify ROIs for the upcoming frame, densely scanning those regions while sparsely sampling non-ROI sections. (b) The Query Prediction module takes the queries of $L$ transformer decoder layers of past frames from a memory buffer to predict the query stacks at time $t$, $Q'^t$, which are then fed to the Mask Generator to predict the final ROI mask.}
 %The History-Aware Query Prediction module using MTM takes the queries of $L$ transformer decoder layers and their corresponding bounding-box outputs over past frames as a memory buffer to predict the queries at time $t$, which are then used to predict the final ROI mask.}
 \vspace{-1.6\baselineskip}
 \label{fig:Structure}
\end{figure*}
\vspace{-3mm}
\section{Related Work}
\label{sec:related_work}
%\vspace{-1mm}
%\subsection{Camera and LiDAR fusion for 3D Detection}
%\vspace{-1mm}
%Modern 3D object detectors can be classified as LiDAR-only, camera-only, or multi-modal fusion methods. LiDAR-only methods achieve high performance using geometric data, either by processing raw points directly \cite{shi2019pointrcnn, zheng2021se, zhang2022not, zhang2023simple} or by discretizing them into pillars \cite{lang2019pointpillars} or voxels \cite{zhou2018voxelnet} before feature extraction \cite{shi2020pv, yin2021center}. Camera-only approaches lift 2D image features into 3D or Bird's-Eye-View (BEV) features \cite{philion2020lift, li2024bevformer} to leverage rich color and texture information. 
\paragraph{Camera and LiDAR Fusion for 3D Detection:}Multi-modal camera and LiDAR fusion leverages complementary strengths of each modality. Early works fuse features from parallel network branches \cite{chen2017multi, ku2018joint} or `paint' 3D points with image semantics \cite{vora2020pointpainting}. Recent algorithms leverage transformer decoders to learn cross-modal interactions, either by aligning and fusing Bird's-Eye-View (BEV) maps from different sensors \cite{liu2023bevfusion}, or by extending the query-based paradigm such as DETR \cite{carion2020end} to 3D \cite{bai2022transfusion, wang2022detr3d, chen2023futr3d}. More recently, CMT \cite{yan2023cross} ingests image and point-cloud tokens directly into a cross-modal transformer that implicitly aligns features via positional embeddings. However, all these methods assume a dense LiDAR scan at every frame without utilizing temporal cues to avoid redundant sensing and excessive power consumption.
\paragraph{Efficient and Adaptive Inference:} Modern perception systems increasingly adopt \textit{adaptive inference} to allocate resources intelligently and avoid redundant computation. Network level methods such as early exiting \cite{bolukbasi2017adaptive, huang2017multi}, dynamic layer skipping \cite{veit2018convolutional, wang2018skipnet}, and channel pruning \cite{hua2019channel, yuan2020s2dnas} enable models to adjust their depth or width according to input difficulty. 
Complementing these methods, adaptive multi-task architectures improve their efficiency by sharing computations across different tasks \cite{shoouri2023efficient}. Adaptive inference techniques \cite{wu2019adaframe, panda2021adamml, meng2021adafuse} efficiently process sequential time-series inputs by training recurrent policies to select only the informative frames \cite{wu2019adaframe, panda2021adamml, meng2021adafuse}. \cite{jang2016categorical} demonstrates an adaptive inference to dynamically adjust image resolution. In multi-modal systems, \cite{yang2022efficient} introduces an adaptive visual and inertial odometry scheme that dynamically disables cameras when a low-power inertial sensing unit provides sufficient motion information. Our framework introduces a new adaptive inference scheme where the LiDAR dynamically changes its scanning density for (non-)ROIs based on the historical context of the scene.
\paragraph{Point Cloud Down-Sampling:}A primary challenge in 3D perception is managing the immense data volume from LiDAR, which makes down-sampling desirable for efficient processing. Classic approaches focus on geometric strategies \cite{cheng2022meta, nezhadarya2020adaptive,wen2023learnable} such as fixed voxel-grid quantization \cite{zhou2018voxelnet} or Farthest Point Sampling to select the subset of points that maximally preserve geometric coverage \cite{qi2017pointnet++}. SampleNet \cite{lang2020samplenet} introduces a differentiable sampler that learns which points are most salient for downstream tasks, while H-PCC \cite{liu2025h} combines predictive encoding with content-adaptive down-sampling to minimize the storage volume of point clouds. All of these methods are post-processing techniques performed \textit{after} obtaining a full LiDAR scan, wasting sensing energy. In contrast, our work controls the LiDAR scan density \textit{before} capture to conserve sensing energy.

\vspace{-3mm}
\section{Proposed Method}
\label{sec:method}
\vspace{-1mm}
%This work introduces an adaptive LiDAR scanning framework that employs a history-aware policy to replace inefficient, uniform $360^\circ$ scans. After defining the limitations of current data acquisition methods (Sec. \ref{sec:Problem formulation})
In this section, we present the adaptive LiDAR-camera fusion model and propose a two-stage architecture for predicting ROIs to guide adaptive LiDAR sensing, as illustrated in Figure \ref{fig:Structure} (a).
%We then describe the history-aware query prediction module, the differentiable mask generation process, the differentiable voxelization technique, and the multi-term loss function that enables end-to-end training.

\vspace{-2.5mm}
\subsection{Framework Overview}
\vspace{-1mm}
\label{sec:Problem formulation}
In the standard sensor fusion for 3D object detection, a LiDAR and synchronized cameras capture scenes at discrete timesteps. Formally, at each timestep $t$, the LiDAR provides a dense point cloud, $P^t = \{\, p_i^{t} \in \mathbb{R}^{D_p} \mid i = 1,\dots,N \}$, where each point $p_i^{t}$ encodes a 3D coordinate and additional features (e.g.\, intensity), and $N$ is the number of points. Simultaneously, the cameras capture a set of $M$ multi-view images, $I^t = \{\, I_m^{t} \mid m = 1,\dots,M \}$. Given these measurements, the objective is to produce a set of 3D bounding boxes, $\mathcal{B}^t=\{b_1^t, \dots, b_B^t\}$, that accurately localize and classify objects in the scene.  Each bounding box $b_i^t$ is parameterized by 7 degrees of freedom (DoF), consisting of its 3D center $C^t=(x_i^t,y_i^t,z_i^t)$, dimensions $(l_i^t,w_i^t,h_i^t)$, and yaw angle $\theta_i^t$, along with an associated class label.

Without loss of generality, we adopt the standard multi-modal 3D detection architecture, as depicted in Figure \ref{fig:Structure} (a). At each timestep $t$, each camera image $I_m^{t}$ is passed through a 2D image backbone to produce a per-view semantic features, $F_{\mathrm{I},m}^t \in \mathbb{R}^{C_{\mathrm{I}}\times H_{\mathrm{I}}\times W_{\mathrm{I}}}$, which are then aggregated into a unified high-level feature map, $F_{\mathrm{I}}^t$. Similarly, the LiDAR point cloud $P^t$ is fed into a LiDAR backbone to extract geometric features, $F_{\mathrm{P}}^t$. These uni-modal features are then merged using a fusion module (either by simple concatenation or via a cross‐modal attention layer) to form the unified multi-modal representation, $F_{fuse}^t$. 

This fused feature $F_{fuse}^t$ serves as the key ($K$) and value ($V$) in a multi-layer transformer decoder. The decoder operates on a fixed set of $N_q$ learnable queries, $Q^t = \{\,q_i^{t} \in \mathbb{R}^D \mid i = 1, \dots, N_q\}$, each representing an object `slot'. At each decoder layer $l$, queries $Q^t_{l}$ first perform multi-head self-attention and then attend to $F_{fuse}^t$ using the standard attention mechanism, defined as: $Attention(Q,K,V)=softmax(QK^\top/ \sqrt D)V$. The output queries from the final layer, $Q^t_{L}$, are then passed to a detection head, composed of feed-forward networks (FFN), to decode each query into 3D bounding-box parameters.
\vspace{-2mm}
\paragraph{LiDAR Scanning Model:} Conventional LiDAR systems operate in a stateless fashion, emitting laser pulses uniformly across all vertical and horizontal angular directions to produce a dense, uniform-angle scan at each frame. This scanning pattern is determined by the sensor’s angular resolution, with $H$ elevation angles (vertical rows) and $W$ azimuthal angles (horizontal columns). Each beam in angular coordinate $(i,j)$ travels a range $r_{i,j}$, and the LiDAR measures it as the sensing data. %and incurs energy approximately proportional to $r^4_{max}$ due to two‐way optical attenuation to cover the maximum range of $r_{max}$. Consequently, the total energy for one full dense scan yields $E_{\mathrm{dense}} = H \times W \times e_{\mathrm{beam}}$, where $e_{\mathrm{beam}} \propto r^{4}_{max}$ is the energy cost per beam. 
In our adaptive LiDAR scan method, we assume that the system can enable/disable any beam with arbitrary elevation and azimuthal scan patterns.
%\vspace{-1mm}
%\begin{equation}
%\label{Lidar_energy}
%\setlength{\abovedisplayskip}{5pt} 
%   E_{\mathrm{scan}}
%= \sum_{i=1}^{H} \sum_{j=1}^{W}
%  E_{\mathrm{beam}}\bigl(r_{i,j}\bigr).
%  \setlength{\belowdisplayskip}{5pt} 
%\end{equation}
%where $E_{\mathrm{beam}}(r) \propto r^{4}$. For analytical simplicity, this can be approximated by a constant per‐beam cost $e_{\mathrm{beam}}$, yielding  $E_{\mathrm{dense}} = H \times W \times e_{\mathrm{beam}}$, which is expended at every timestep regardless of scene dynamics. In practice, such uniform acquisition leads to substantial energy waste, as the majority of pulses redundantly scan static background regions that seldom change between consecutive frames.
\vspace{-2mm}
\paragraph{Range Image Representation:} We represent the LiDAR's acquisition pattern using a 2D format known as a \textit{range image}. This representation is common in methods that process the LiDAR data using 2D convolutional models \cite{ milioto2019rangenet++}. Formally, a range image $R^t \in \mathbb{R}^{H \times W}$ is a projection where each pixel %$(u,v)$
corresponds to a unique beam direction defined by an azimuth angle $\theta$ and an elevation angle $\phi$. The projection from a 3D point $p_i^t=(x_i,y_i,z_i)$ to its spherical coordinates $(r_i, \theta_i, \phi_i)$ is defined as:
\vspace{-1mm}
\begingroup
  % tighten skips for this display only
  \setlength{\abovedisplayskip}{4pt}
  \setlength{\abovedisplayshortskip}{4pt}
  \setlength{\belowdisplayskip}{2pt}
  \setlength{\belowdisplayshortskip}{2pt}
\begin{gather}
\scalebox{0.9}{$
   \begin{split}
    r_i = \sqrt{x_i^2 + y_i^2 + z_i^2},  \:
    \phi_i = \arcsin\!\bigl(\tfrac{z_i}{r_i}\bigr),  \: \theta_i = \operatorname{atan2}(y_i, x_i).\\
   \end{split}$}
\end{gather}
\endgroup
%\vspace{-2mm}
These spherical coordinates are then discretized to map to pixel coordinates $(u,v)$ based on the sensor's vertical and horizontal field-of-view, such that:
\vspace{-1mm}
\begin{equation}
\label{Lidar_energy}
\setlength{\abovedisplayskip}{4pt} 
  u
= \left\lfloor
    (\phi_i - \phi_{\min})/{\Delta \phi}
  \right\rfloor,  \; \; \; \; v
= \left\lfloor
    (\theta_i - \theta_{\min})/{\Delta \theta}
  \right\rfloor,
  \setlength{\belowdisplayskip}{2pt} 
\end{equation}
where $\Delta \phi=(\phi_{\max}-\phi_{\min})/H$ and $\Delta \theta=(\theta_{\max}-\theta_{\min})/W$. Each pixel $(u,v)$ of a range image $R^t$ records the minimum $r_i$ of all LiDAR points that belong to that pixel coordinate. By representing the scan pattern as this range image, the problem of adaptive scanning is formulated as a task of learning a binary mask that dictates which range image pixels require LiDAR sensing at each timestep.
\vspace{-2mm}
\subsection{Two‐Stage History‐Aware Adaptive Scanning}\label{sec:Two-Stage}
\vspace{-1mm}
To estimate the optimal scanning mask $M^t$, we propose a two-stage Prediction Policy model. It leverages temporal context to make informed predictive decisions, and it can be plugged into any multi-modal fusion 3D detectors. The first stage, a \emph{History-Aware Query Prediction} module, forecasts the state of objects in the upcoming frame solely based on past observations. The second stage, a \emph{Differentiable Mask Generator}, then translates these object-centric predictions into a mask pattern in the range image space, as shown in Figure \ref{fig:Structure} (b).
\vspace{-2mm}
\paragraph{History-Aware Query Prediction using Motion-guided Temporal Module:}
Our model predicts queries for the current timestep $t$, $Q'^t$, using a historical buffer containing object query sets from the past $T$ frames, denoted $\{Q^{t-T}, \dots, Q^{t-1}\}$, along with their predicted 3D bounding box centers $\{C^{t-T}, \dots, C^{t-1}\}$ and velocities $\{V^{t-T}, \dots, V^{t-1}\}$. Our architecture builds upon the core principles of the Motion-guided Temporal Module (MTM) from QTNet \cite{hou2023query}. While the original MTM fuses past queries with current frame data to `refine' current detections, we repurpose its core concept into an autoregressive framework for a purely predictive task where no current frame sensor data is used/available.

As illustrated in Figure \ref{fig:Structure} (b), for each layer $l\in\{1, \dots, L\}$ in the transformer decoder, we sequentially unroll $T-1$ previous time steps through the historical buffer. At each step $\tau$ (from $\tau=T$ down to $2$), the module takes a pair of query embeddings $Q_l^{t-\tau}$ and $Q_l^{t-\tau+1}$ along with their corresponding bounding box centers $C^{t-\tau}$ and $C^{t-\tau+1}$ as input to predict the queries for the subsequent timestep. To ensure spatial consistency, we first align the bounding box center $C^{t-\tau}$ into the coordinate frame of time $t-\tau+1$. Using the known world‑to‑sensor rotations
$R_w^{t-\tau+1}$ and $R_{t-\tau}$, we compute the relative rotation as $R_{t-\tau}^{t-\tau+1}=R_w^{t-\tau+1}(R_w^{t-\tau})^{-1}$. Applying a constant-velocity motion model, the aligned historical center is then calculated as $C'^{t-\tau}=(C^{t-\tau}+V^{t-\tau}\Delta t)(R_{t-\tau}^{t-\tau+1})^{\top}$, where $\Delta t$ is the inter‐frame interval.

Using the aligned centers, a geometry-guided attention map is generated to associate objects across time. First, a cost matrix is defined based on the pairwise $L_2$ norm (Euclidean distance): $O_{t-\tau}^{t-\tau+1}=||C^{t-\tau+1}-C'^{t-\tau}||_{2}$.
Next, we construct a guided mask $G_{t-\tau}^{t-\tau+1}$ to penalize improbable matches between different object classes or those separated by more than a distance threshold $\gamma$:
\begingroup
  \setlength{\abovedisplayskip}{1pt}   % space above
  \setlength{\belowdisplayskip}{1pt}   % space below
\begin{equation}
G_{t-\tau}^{t-\tau+1} =
\begin{cases}
0, & O_{t-\tau}^{t-\tau+1} \le \gamma \;and\; s_{t-\tau} = s_{t-\tau+1}\\[2pt]
c_{m}, & O_{t-\tau}^{t-\tau+1} > \gamma \;or\; s_{t-\tau} \neq s_{t-\tau+1},
\end{cases}
\end{equation}
\endgroup
where $s_{t}$ indicates the object category at time $t$ and $c_{m}$ is a large constant (e.g., $10^8$). The attention map is then defined as $A_{t-\tau}^{t-\tau+1}=softmax(-O_{t-\tau}^{t-\tau+1}-G_{t-\tau}^{t-\tau+1})$, as shown in Figure \ref{fig:Structure} (b) (MTM module).

This attention map is used to aggregate historical query features into a context vector, formulated as $F^{t-\tau}_l=\Phi_2(A_{t-\tau}^{t-\tau+1}\Phi_1(Q_l^{t-\tau}))$, where $\Phi_1$ and $\Phi_2$ denote two linear layers.
 After aggregation, $F^{t-\tau}_l$ and $Q_l^{t-\tau+1}$ are combined to produce a provisional Query Prediction, $Q_l'^{t-\tau+2}$, calculated as $Q_l'^{t-\tau+2}=FFN(Norm(Q_l^{t-\tau+1}+F^{t-\tau}_l))$, where $Norm$ represents the layer normalization. This provisional query is the output of the MTM module (Figure \ref{fig:Structure} (b)). 
 
The provisional $Q_l'^{t-\tau+2}$ is fused with its corresponding historical embedding ($Q_l^{t-\tau+2}$) by concatenating them and then applying a pointwise convolutional projection. This produces an updated $Q_l^{t-\tau+2}$, which serves as the input for the next iteration as shown in Figure \ref{fig:Structure} (b). This alignment-attention-aggregation-fusion procedure is repeated sequentially for each time step in the historical buffer at every decoder layer. Ultimately, this process yields the final predicted query set $Q'^t$ for all decoder layers that carries both learned priors and motion-guided history.
\vspace{-2mm}
\paragraph{Differentiable Mask Generator:} To generate an adaptive LiDAR scan, we feed the predicted query stack from all $L$ decoder layers, $Q'^t\in \mathbb{R}^{L\times N_q \times D}$, to the Differentiable Mask Generator. This module translates these object-centric representations into a LiDAR scanning policy represented by a spatial logits map, $Z^t\in \mathbb{R}^{H_B \times W_B \times 2}$, for the LiDAR range image view that consists of  $H_B \times W_B$ `block's. Each spatial location $(u,v)$ corresponds to one angular block in the LiDAR's range image view and encodes the probability of performing either a `full' (dense) or a `sparse' scan for that block. Formally, the logits map is computed as $Z^{t} = \mathcal{F}_{\mathrm{Head}}\bigl(\mathcal{F}_{\mathrm{Enc}}(Q'^t)\bigr)$. The encoder $\mathcal{F}_{\mathrm{Enc}}$ consists of depthwise convolution residual blocks that operate on each channel independently, followed by standard residual blocks, and a final $1\times1$ convolution to project those channels into two class logits (full vs. sparse scan). The subsequent module, $\mathcal{F}_{\mathrm{Head}}$ applies an adaptive pooling layer and an FFN to generate the final logits. A binary scan mask, $M^{t}$, is then drawn from $Z^{t}$ via the Gumbel-Softmax reparameterization (instead of non-differentiable Bernoulli draws) with a temperature ($\tau$)‐controlled softmax over Gumbel‐perturbed logits: $M^{t} \sim \mathrm{GumbelSoftmax}\bigl(Z^{t};\tau\bigr)$. Specifically, for each block with spatial location $(u,v)$, the probability of scanning mode $k\in \{full,sparse\}$ is obtained by:
%\vspace{-5mm}
\begingroup
  % tighten skips for this equation only
  \setlength{\abovedisplayskip}{2pt}
  \setlength{\abovedisplayshortskip}{2pt}
  \setlength{\belowdisplayskip}{2pt}
  \setlength{\belowdisplayshortskip}{2pt}
\begin{equation}
\label{Gumbel}
\tilde M^t_{k}(u,v)
=
\frac{
  \exp\!\bigl((Z^t_{k}(u,v) + g_{k})/\tau\bigr)
}{
  \displaystyle\sum_{j\in\{full,sparse\}}
    \exp\!\bigl((Z^t_{j}(u,v) + g_{j})/\tau\bigr)
}\,,
\end{equation}
\endgroup
%\vspace{-3mm}
where $g_k=-log(-logU_k)$ is a standard Gumbel distribution with $U \sim \mathrm{Uniform}(0,1)$ and $\tau$ controls how sharply $\tilde M^t$ approximates a one‐hot vector. The binary mask $M^t$ is only used for forward pass while the soft probability $\tilde M^t$ is used to propagate gradients during training. 

During inference, we generate the final adaptive LiDAR scanning pattern using a two-level sampling strategy. For blocks marked for full scans ($M^t(u,v)=1$), we activate all LiDAR points. For blocks with sparse scans ($M^t(u,v)=0$), we employ probabilistic subsampling based on the corresponding soft probability $\tilde M^t_{full}(u,v)$. This soft probability is first quantized to a predefined sparsity level (e.g., $6.25\%$, $12.5\%$, etc.). Then, a final sparse scanning pattern is generated by performing stochastic Bernoulli sampling for each beam within the block, using the quantized probability as the sampling rate. The resulting sparse point cloud is then fed into the downstream detection network. By adjusting the quantized sparsity level, the model can make a tradeoff between LiDAR efficiency and object detection accuracy.
\vspace{-3mm}
\subsection{Differentiable Voxelization}
\label{Sec:Voxelization}
\vspace{-1.5mm}
Conventional voxelization \cite{zhou2018voxelnet} is inherently non-differentiable, preventing the gradient of the final detection loss from propagating through it to the Mask Generator. To address this, we propose a heuristic differentiable voxelization method that approximates gradients through nearest-neighbor assignments. In the forward pass, we apply standard voxelization \cite{zhou2018voxelnet} to the sparse LiDAR points, $P_{\mathrm{s}}^{t}\in \mathbb{R}^{N_{s} \times D_p}$, producing a voxel tensor $V\in \mathbb{R}^{M_v\times K_v \times D_p}$, where $M_v$ is the number of occupied voxels, $K_v$ is the maximum points per voxel, and $D_p$ is the per-point feature dimension. During back-propagation, we approximate the gradient $\frac{\partial Loss}{\partial P_{s}}$ from the voxel gradient $\frac{\partial Loss}{\partial V}$ by assigning each point to its nearest voxel center: $\frac{\partial Loss}{\partial P_{s, i}}
\;\approx\;
\alpha \,\frac{\partial Loss}{\partial V_{\mathrm{NN(i)}}}$, where 
$NN(i)$ is the index of the voxel whose center is closest to point $P_{s, i}$ and $\alpha$ is a scaling factor that stabilizes training. This simple yet effective heuristic restores gradient propagation through the voxelization step, making the entire pipeline end-to-end trainable.
\vspace{-3mm}
\subsection{Loss Function for Training}
\label{Sec:loss}
\vspace{-1.5mm}
Our framework is trained end‐to‐end by minimizing a single composite loss averaged across $T-1$ consecutive frames:
%\vspace{-3mm}
\begingroup
  % tighten skips for this equation only
  \setlength{\abovedisplayskip}{2pt}
  \setlength{\abovedisplayshortskip}{2pt}
  \setlength{\belowdisplayskip}{2pt}
  \setlength{\belowdisplayshortskip}{2pt}
\begin{gather}
\scalebox{0.86}{$
   \begin{split}
\mathcal{L}
=\frac{1}{T-1} \sum_{t=2}^{T}
\Bigl(
  \mathcal{L}_{3D}^t
  + \lambda_1\,\mathcal{L}_{\mathrm{distill}}^t
  + \lambda_2\,\mathcal{L}_{\mathrm{mask}}^t
  + \lambda_3\,\mathcal{L}_{\mathrm{CVaR}}^t
\Bigr).
   \end{split}$}
\end{gather}
\endgroup
The 3D object detection loss $\mathcal{L}_{3D}^t$ combines classification and regression terms to compare the predicted bounding boxes ($\mathcal{B}^t$) with the ground-truth ($\mathcal{B}_{gt}^t$). To enforce temporal consistency, the distillation loss aligns the predicted queries with the reference queries $Q_{ref}^{t}$ generated by the same fusion model on the full-dense point clouds: $\mathcal{L}_{\mathrm{distill}}
= \bigl\|{Q}'^{\,t} - Q_{ref}^{t}\bigr\|_{1}$. To guide the Mask Generator, we define the mask loss, $\mathcal{L}_{\mathrm{mask}}^t$. We first project the ground-truth bounding box positions onto the LiDAR range image to produce a binary guidance mask, $M_{GT}^t$, then compute a per-pixel Focal Loss ($\mathrm{FL}$) \cite{lin2017focal} to handle the class imbalance between foreground (object) and background pixels:
%\vspace{-2.5mm}
\begingroup
  % tighten skips for this equation only
  \setlength{\abovedisplayskip}{4pt}
  \setlength{\abovedisplayshortskip}{4pt}
  \setlength{\belowdisplayskip}{4pt}
  \setlength{\belowdisplayshortskip}{4pt}
\begin{gather}
\scalebox{0.88}{$
   \begin{split}
   \mathcal{L}_{\mathrm{mask}}^t
= \frac{1}{H_B\,W_B}
\sum_{u=1}^{H_B}\sum_{v=1}^{W_B}
\mathrm{FL}\!\bigl({M}_{(u,v)}^t,{M^t_{GT}}_{(u,v)} \bigr).\\
   \end{split}$}
   \setlength{\belowdisplayskip}{-3mm}
\end{gather}
\endgroup
Finally, to improve the robustness of the Mask Generator for small yet critical objects (e.g.\, pedestrians), we introduce a Conditional Value at Risk (CVaR) loss that focuses on worst-case prediction errors. For each sample, we extract the per-pixel mask losses $\{\ell_{k}\}_{k=1}^{N_{small}}$ from $\mathcal{L}_{\mathrm{mask}}^t$ for regions corresponding to small objects and sort them in descending order. We define the `Value at Risk', $m^*$, as the loss of the $\lceil(1 - \beta)\,N_{small}\rceil$-th worst sample, such that: $m^* = \ell_{(\lceil(1 - \beta)\,N_{small}\rceil)}$. Then, we formulate the CVaR loss: 
\vspace{-2mm}
\begingroup
  % tighten skips for this equation only
  \setlength{\abovedisplayskip}{4pt}
  \setlength{\abovedisplayshortskip}{4pt}
  \setlength{\belowdisplayskip}{4pt}
  \setlength{\belowdisplayshortskip}{4pt}
\begin{gather}
\scalebox{0.89}{$
   \begin{split}
\mathcal{L}^t_{\mathrm{CVaR}}
= m^*
\;+\;
\frac{1}{\beta\,N_{small}}
\sum_{k=1}^{N_{small}}
\max\bigl(\ell_{k} - m^*,\,0\bigr).
   \end{split}$}
   \setlength{\belowdisplayskip}{-1mm}
\end{gather}
\endgroup
By penalizing the average of the worst $\beta$-fraction of losses, CVaR drives the network to improve performance on the most challenging and small-object regions.

\begin{table*}[t]
  \centering
  \footnotesize
   \renewcommand{\arraystretch}{0.88} % slightly taller rows
  \caption{Performance comparison using next-frame prediction on nuScenes validation set. Scan sparsity = percentage of empty pixels.}
   \label{tab:Nus-Nextframe}
  \vspace{-3mm}
  \begin{tabular}{@{}l|c|l|c|c|c@{}}
    \toprule
    \textbf{Model}   & \textbf{Image Res.} & \textbf{Backbone}     & \textbf{mAP\% $\uparrow$\,} & \textbf{NDS\% $\uparrow$\,} & \textbf{Scan Sparsity\% $\uparrow$\,} \\
    \midrule
    AutoAlignV2 \cite{chen2022deformable}   & 1600×900  & CSPNet \& VoxelNet      & 64.4  & 69.5   & 0   \\
    FUTR3D \cite{chen2023futr3d}   & 1600×900  & VoVNet \&  VoxelNet   & 64.5   & 68.3   & 0   \\
    UVTR \cite{li2022unifying}  & 1600×900  & R50 \& VoxelNet      & 65.4   & 70.2  & 0   \\
    MSMDFusion \cite{jiao2023msmdfusion}   & 448×800 & R50 \& VoxelNet      & 66.9   & 68.9   & 0   \\
    MVP \cite{yin2021multimodal}  & 1600×900  & DLA34 \& VoxelNet      & 67.1   & 70.8   & 0    \\
    TransFusion \cite{bai2022transfusion}   & 448×800  & R50 \& VoxelNet & 67.5  & 71.3   & 0   \\
    BEVFusion \cite{liang2022bevfusion}   & 448×800  & Swin-Tiny \& VoxelNet   & 67.9   & 71.0   & 0   \\
    BEVFusion \cite{liu2023bevfusion}   & 448×800  & Swin-Tiny \& VoxelNet   & 68.5   & 71.4   & 0   \\
    DeepInteraction \cite{yang2022deepinteraction}   & 1600×900 & R50 \& VoxelNet      & 69.9   & 72.6   & 0   \\
    \midrule  % <-- extra line here
    CMT \cite{yan2023cross}  & 800×320 & R50 \& VoxelNet      & 67.9   & 70.7   & 0   \\
    \rowcolor{gray!15}
    +Adaptive Scan  & 800×320 & R50 \& VoxelNet    & 67.5   & 70.6   & \textbf{66.0}   \\
    CMT  & 1600×640 & VoVNet \& VoxelNet           & 70.3   & 72.9   & 0    \\
    \rowcolor{gray!15}
    +Adaptive Scan   & 1600×640 & VoVNet \& VoxelNet   & \textbf{71.0}   & \textbf{73.1}   & \textbf{66.8}   \\
    \bottomrule
  \end{tabular}
  \vspace{-3mm}
\end{table*}
\begin{table*}[t]
  \centering
  \footnotesize
  \renewcommand{\arraystretch}{0.8} % slightly taller rows
  \caption{Performance comparison using next-frame prediction on Lyft validation set. `Other Veh.' is other vehicle.}
  \label{tab:lyft-nextframe}
  \vspace{-3mm}
  \begin{tabular}{@{}l|c|c|c|c c c c c c c@{}}
    \toprule
    \textbf{Model} & \textbf{mAP\% $\uparrow$\,} & \textbf{w-mAP\% $\uparrow$\,} & \textbf{Sparsity\% $\uparrow$} & \textbf{Car} & \textbf{Other Veh.} & \textbf{Bus} & \textbf{Truck} & \textbf{Motorcycle} & \textbf{Bicycle} & \textbf{Pedestrian}\\
    \midrule

    CMT  & 48.4 & 79.8          & 0   & 85.5   & 75.7 & 50.2& 54.7 & 31.4 &   21.1 & 20.5  \\
    \rowcolor{gray!15}
    +Adaptive Scan   & 48.4 & \textbf{80.4}  & \textbf{68.7}   & \textbf{87.1}   & \textbf{77.5} & 46.6 & 50.2 & 23.0 & \textbf{29.3} & \textbf{25.2}   \\
    \bottomrule
  \end{tabular}
  \vspace{-5mm}
\end{table*}
\begin{table}[!htbp]
  \centering
  \footnotesize
  \renewcommand{\arraystretch}{0.9}
  \caption{Performance comparison using next-frame prediction on nuScenes test set. $^{*}$ denotes our reproduced results.}
  \vspace{-3mm}
  \label{Nus-nextframe-test}
  \resizebox{1\columnwidth}{!}{%
    \begin{tabular}{@{}c|c|c|c@{}}
      \toprule
      \textbf{Model}
      & \textbf{mAP\% $\uparrow$}
        & \textbf{NDS\% $\uparrow$}
        & \textbf{Sparsity\% $\uparrow$}
        \\
      \midrule
          PointPainting\cite{vora2020pointpainting}    &  54.1  &  61.0         &   0 \\
          PointAugmenting \cite{wang2021pointaugmenting} & 66.8 & 71.1 & 0 \\
          MVP \cite{yin2021multimodal} & 66.8 & 70.5 & 0 \\
          UVTR \cite{li2022unifying} & 67.1 & 71.1 & 0 \\
          FusionPainting \cite{xu2021fusionpainting} & 68.1 & 71.6 & 0\\
          TransFusion \cite{bai2022transfusion} & 68.9 & 71.7 & 0 \\
          BEVFusion \cite{liu2023bevfusion} & 70.2 & 72.9 & 0 \\
          \midrule  % <-- extra line here
         CMT$^{*}$ (800×320)  & 68.6 & 71.2 & 0\\
          \rowcolor{gray!15}
         +Adaptive Scane & 68.3     & 71.0   & \textbf{63.8}  \\
         CMT$^{*}$ (1600×640)  & 70.4 & 73.0 & 0\\
         \rowcolor{gray!15}
         +Adaptive Scane & 70.0     & 72.9   & \textbf{64.2}  \\
      \bottomrule
    \end{tabular}%
  }
  \vspace{-6mm}
\end{table}
\begin{table*}[!htpb]
  \centering
  \begin{minipage}{0.48\textwidth}
    \centering
    \footnotesize
    \renewcommand{\arraystretch}{0.88}
    \caption{Evaluation of entire sequence prediction on nuScenes validation set}    \vspace{-3mm}
    \label{Nus-temporal}
    \resizebox{\linewidth}{!}{%
      \begin{tabular}{@{}l|c|c|c|c@{}}
        \toprule
        \textbf{Model}
          & \textbf{mAP\% $\uparrow$}
          & \textbf{NDS\% $\uparrow$}
          & \textbf{Sparsity\% $\uparrow$}
          & \textbf{GFLOPs} \\
        \midrule
        CMT (800×320)              & 67.9   & 70.7   & 0          & 769.4   \\
        \rowcolor{gray!15}
        +Adaptive Scan      & 67.5   & 70.6   & \textbf{65.3} & \textbf{742.2}  \\
        CMT (1600×640)             & 70.3   & 72.9   & 0          & 3260.9 \\
        \rowcolor{gray!15}
        +Adaptive Scan      & \textbf{71.0}      & \textbf{73.1}      & \textbf{66.4}            & \textbf{3175.2}      \\
        \bottomrule
      \end{tabular}%
      
    }
  \end{minipage}\hfill
  \begin{minipage}{0.48\textwidth}
    \centering
    \footnotesize
    \renewcommand{\arraystretch}{0.88}
    \caption{Evaluation of entire sequence prediction on Lyft validation set}
    \vspace{-3mm}
    \label{Lyft-temporal}
    \resizebox{\linewidth}{!}{%
      \begin{tabular}{@{}l|c|c|c|c@{}}
        \toprule
        \textbf{Model}
          & \textbf{mAP\% $\uparrow$}
          & \textbf{w-mAP\% $\uparrow$}
          & \textbf{Sparsity\% $\uparrow$}
          & \textbf{GFLOPs} \\
        \midrule
        CMT              & 48.4   & 79.8   & 0          & 3351.2   \\
        \rowcolor{gray!15}
        +Adaptive Scan      &  \textbf{48.6}  &  \textbf{80.5}  & \textbf{67.2} & \textbf{3289.6}  \\
        \bottomrule
      \end{tabular}%
    }
  \end{minipage}
  \vspace{-6mm}
\end{table*}
\vspace{-2mm}
\section{Experiments}
\label{experiments}
\subsection{Experimental Setup}
\vspace{-1mm}
\paragraph{Dataset:} We evaluate our framework on the large-scale nuScenes \cite{caesar2020nuscenes} and Lyft datasets \cite{lyft2019}. NuScenes comprises $1000$ scenes split into $700/150/150$ for training/validation/testing. Each $20$ second scene is captured with a full sensor suite, including a $32$-beam LiDAR spinning at $20Hz$ and six cameras providing $360^\circ$ coverage, with the 3D bounding box annotations for $10$ distinct object categories provided at $2Hz$ on keyframes. For our experiments, we follow the official protocol and evaluate our model at $2Hz$ frequency to ensure a fair comparison with existing methods. The Lyft dataset offers a higher annotation rate of $5Hz$ for 3D bounding boxes, providing finer temporal granularity for our history‐aware Query Predictions. It is collected in a complex urban environment with a $64$-beam roof LiDAR and six cameras, containing over $10000/3400$ samples for training/validation.
\vspace{-1.5mm}
\paragraph{Metrics:} For nuScenes, we report performance using the official evaluation metrics: mean Average Precision (mAP) over a set of distance thresholds, and the primary nuScenes Detection Score (NDS). For Lyft dataset, we report the standard Average Precision (AP) for object categories and the overall mAP. Furthermore, to account for the significant class imbalance in the Lyft dataset, we introduce a weighted-mAP (w-mAP) that weights the AP of each class by the frequency of its instances so that the metric is not dominated by relatively rare object types such as motorcycles.
\vspace{-1.5mm}
\paragraph{Implementation Details:} Our adaptive LiDAR scanning module is designed as a plug‐in that seamlessly integrates with any query‐based camera‐LiDAR fusion architecture. For our experiments, we adopt CMT \cite{yan2023cross} as the base fusion model for its effective performance and stable convergence, which contains six layers of transformer decoder with $900$ object queries. We use a ResNet \cite{he2016deep} (low image resolution) or VoVNet \cite{lee2020centermask} (high image resolution) model as image backbone, and VoxelNet \cite{zhou2018voxelnet} as the LiDAR backbone.
For temporal context, our history‐aware prediction module uses a buffer of $T=4$ frames for both datasets. %On nuScenes, we follow the standard protocol data inputs, similar to \cite{yan2023cross,hou2023query}. The point cloud range is set to $[-54\,\mathrm{m},\,54\,\mathrm{m}]$ for the $X$ and $Y$ axes and $[-5\,\mathrm{m},\,3\,\mathrm{m}]$ for the $Z$ axis, with a voxel size of $(0.075\,\mathrm{m},\;0.075\,\mathrm{m},\;0.1\,\mathrm{m})$. For the Lyft dataset, we use a larger range of $[-100\,\mathrm{m},\,100\,\mathrm{m}]$ for $X/Y$ and a voxel size of $(0.125\,\mathrm{m},\;0.125\,\mathrm{m},\;0.2\,\mathrm{m})$.

We employ a three-stage training strategy to ensure stable convergence. First, we train the Query Predictor in isolation by attaching the pre-trained (frozen) CMT and supervising it to generate accurate 3D bounding boxes from the Query Predictor output at future time $t$ predicted based on the buffered queries obtained at $t-1, \dots, t-T$. Then, we train the CMT to operate on sparse LiDAR inputs by initializing the Query Predictor with weights from the first stage and integrating it with the Mask Generator that sparsifies the LiDAR input. Although voxelization remains non‐differentiable at this point, this step solidifies the ability of the fusion model to process sparse LiDAR inputs. Finally, the entire pipeline is fine-tuned end-to-end by replacing the standard voxelizer with the proposed differentiable voxelization layer, enabling gradient propagation from the final detection loss back to the Mask Generator. 
\vspace{-3mm}
\subsection{Comparison to the State of the Art}
\vspace{-1mm}
We compare our adaptive scanning approach in two scenarios: `next‑frame prediction' and `entire‑sequence prediction'. In the next-frame prediction case, $T$ past frames in the query queue are obtained by full scan to perform the adaptive scan for the next frame. On the other hand, the entire-sequence prediction always uses adaptive scan for all frames including the $T$ recent past frames in the query buffer. 

For the next-frame prediction setting on the nuScenes validation set, we achieve $66.0\%$ LiDAR sparsity at 800×320 resolution while incurring marginal performance degradation of $0.59\%$ in mAP and $0.14\%$ in NDS, as shown in Table \ref{tab:Nus-Nextframe}. Higher image resolution (1600×640) yields $66.8\%$ sparsity while actually improving mAP by $1.0\%$ and NDS by $0.27\%$. On the nuScenes test split, our model achieves sparsity level of $63.8\%$ (low resolution) and $64.2\%$ (high resolution) with competitive performances (Table. \ref{Nus-nextframe-test}). Note that the competitive performance is attained despite the dataset’s low ($2$Hz) frame rate, which makes the next frame prediction challenging. On the higher frame rate (5Hz) Lyft dataset (Table \ref{tab:lyft-nextframe}), we attain $68.7\%$ sparsity without sacrificing mAP and enhancing w-mAP by $0.75\%$. By selectively removing unnecessary background points, our approach surpasses the full-scan baseline in w-mAP, highlighting the benefits of reducing redundant sensor data.
\vspace{-2.5mm}
\begin{figure}[htbp]
  \centering
  % left half
  \begin{minipage}[t]{0.48\textwidth}
    \centering
    \includegraphics[
      width=\linewidth,
      height=0.45\textwidth,
      keepaspectratio
    ]{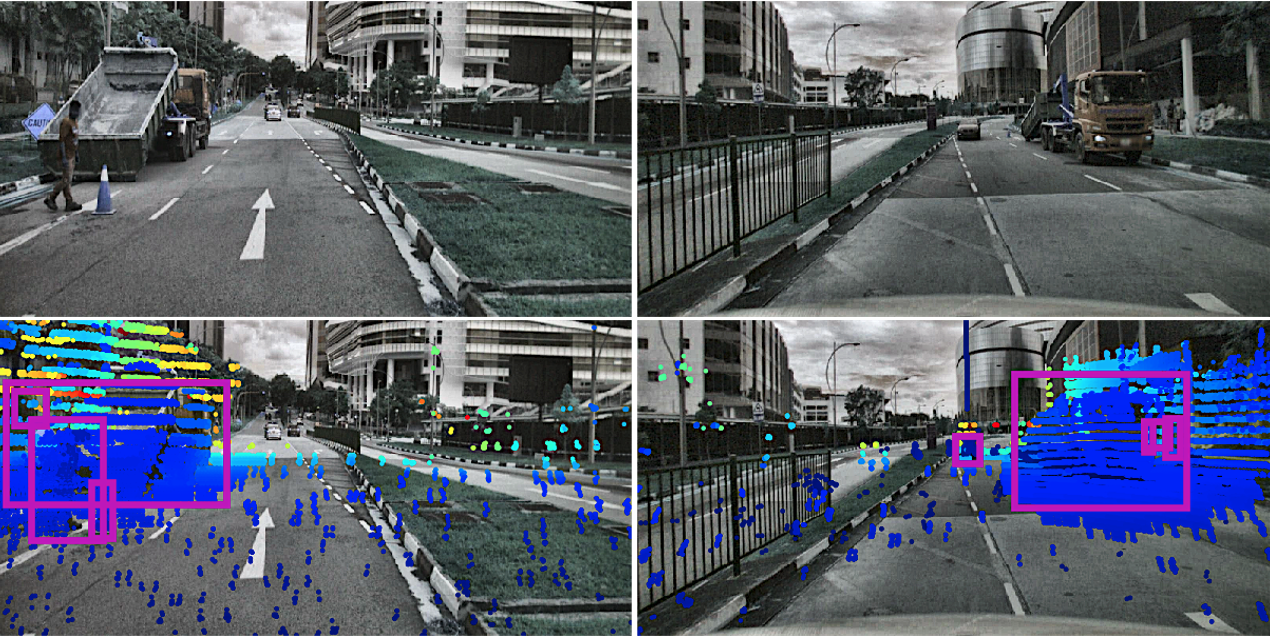}
    \\[\smallskipamount]
  \end{minipage}

  % right half
  \vspace{-2mm}
  \caption{%
 Top: original images from two nuScenes validation examples. Bottom: corresponding selected LiDAR points that are densely scanned inside each ROI and sparsely elsewhere, with point colors encoding distance from the ego-vehicle (low resolution).}
  \vspace{-4mm}
  \label{fig:Visualization}
\end{figure}
%\vspace{-1mm} 

We also evaluate the `entire-sequence prediction' task, where only the initial buffer frames ($T=4$) use a full scan and the remaining frames always use the proposed adaptive LiDAR scanning. On the nuScenes validation dataset (excluding the initial buffer), our method maintains sparsity rates of $65.3\%$ and $66.4\%$ at low and high resolution, maintaining comparable or even improved detection accuracy (Table \ref{Nus-temporal}). Similarly, on the Lyft dataset (Table \ref{Lyft-temporal}), we achieve $67.2\%$ sparsity and boost performance by $0.41\%$ in mAP and $0.88\%$ in w-mAP. Crucially, by scanning only a sparse subset of beams, far fewer points are fed into the LiDAR backbone, which reduces the overall model GFLOPs. Specifically, our method reduces the LiDAR backbone’s GFLOPs by $3.54\%$ (low resolution) and $2.63\%$ (high resolution) on nuScenes, and by $1.84\%$ on Lyft. This demonstrates that our framework not only avoids power consumption overhead in LiDAR sensing but also actively reduces the computational demands of the perception model despite that we introduced additional modules such as the Query Predictor and Mask Generator. Figure \ref{fig:Visualization} provides visualization examples of our adaptive scanning strategy on two frames from the nuScenes validation set, illustrating how the model densely scans regions containing objects and their immediate surroundings (ROIs), while significantly reducing scanning density in non-ROI areas.
\vspace{-2mm}
\subsection{Ablation Study}
\begin{figure}[htbp]
  \centering
  % left half
  \begin{minipage}[t]{0.45\textwidth}
    \centering
    \includegraphics[
      width=\linewidth,
      height=0.45\textwidth,
      keepaspectratio
    ]{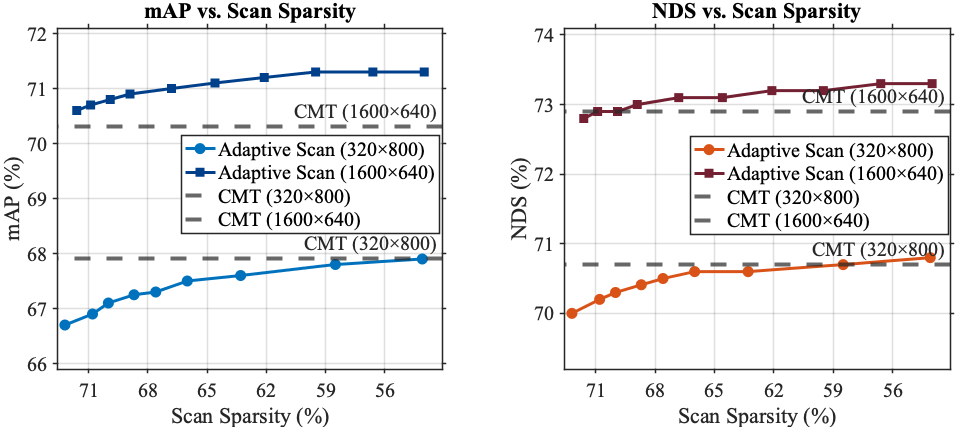}
    \\[\smallskipamount]
  \end{minipage}

  % right half
  \vspace{-2mm}
  \caption{%
 Ablation study on comparing performance under different LiDAR levels on nuScenes validation set.}
  \vspace{-5mm}
  \label{fig:comparison}
\end{figure}

\begin{figure}[htbp]
  \centering
  % left half
  \begin{minipage}[t]{0.48\textwidth}
    \centering
    \includegraphics[
      width=\linewidth,
      height=0.48\textwidth,
      keepaspectratio
    ]{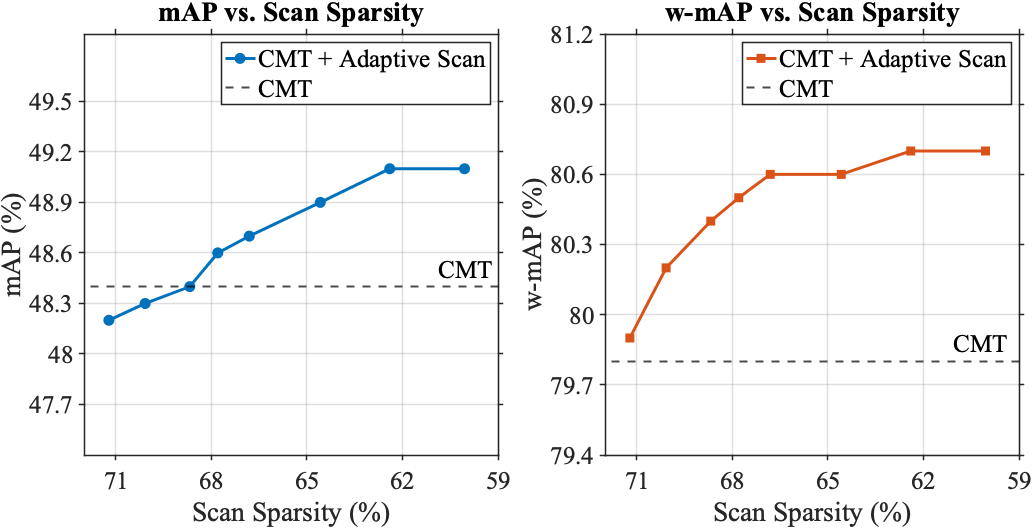}
    \\[\smallskipamount]
  \end{minipage}

  % right half
  \vspace{-2mm}
  \caption{%
 Sparsity ratios vs. Lyft validation set performance 
  }
  \vspace{-4mm}
  \label{fig:comparison-lyft}
\end{figure}

\vspace{-1mm}
\paragraph{Performance vs. Sparsity:} We analyze the trade-off between LiDAR scan sparsity and performance for the `next-frame prediction' case on both nuScenes and Lyft. As illustrated in Figures \ref{fig:comparison} and \ref{fig:comparison-lyft}, reducing scan sparsity (i.e., more points retained) yields higher mAP and NDS (for nuScenes) or w‑mAP (for Lyft). On high‑resolution nuScenes validation, our adaptive scan outperforms the full‑scan CMT in mAP once sparsity drops below about $71.6\%$ and surpasses its NDS at  $70.9\%$ sparsity. On Lyft, we match or exceed the full‑scan CMT’s mAP at  $68.7\%$ sparsity and beat its w-mAP at  $71.2\%$. 
\vspace{-2.5mm}
\vspace{-1mm}
%\vspace{-mm}
\paragraph{Impact of Buffer Length:} We study how the length of the temporal buffer influences accuracy and sparsity on the nuScenes validation set for the `next‑frame prediction' case (low resolution). Table \ref{tab:nuScenes_ablation_frames} shows that as we increase the number of buffer frames, performance steadily improves, while scan sparsity naturally declines. Leveraging more historical context enables the model to anticipate objects that appear further away or re-enter the scene, but it also requires additional computation. However, even with a buffer of $T=4$ frames, the overall processing cost remains lower than the full-scan CMT thanks to the sparse LiDAR input.

\begin{table}[!h]
  \centering
  \begin{minipage}{0.4\textwidth}
    \centering
    \footnotesize
    \renewcommand{\arraystretch}{0.9}
    \vspace{-2mm}
    \caption{Buffer size ($T$) vs. nuScenes performance}
    \vspace{-3mm}
    \label{tab:nuScenes_ablation_frames}
    \resizebox{\linewidth}{!}{%
      \begin{tabular}{@{}c|c|c|c|c@{}}
        \toprule
        
        \textbf{Buffer size} ($T$)& \textbf{mAP\% $\uparrow$}
          & \textbf{NDS\% $\uparrow$}
          & \textbf{Sparsity\% $\uparrow$}
          & \textbf{GFLOPs} \\
        \midrule
        2            &  66.7  & 70.0   &    69.9       &  736.3  \\
        3            &  67.0   & 70.2   &    67.4       &  738.7  \\
        4            & \textbf{67.5}   & \textbf{70.6}   & 66.0   &  742.2  \\
        \bottomrule
      \end{tabular}%
    }
  \end{minipage}
\vspace{-5mm}
\end{table}

\paragraph{Impact of Loss Terms:} We investigate the effects of the proposed losses, $\mathcal{L}_{\mathrm{CVaR}}$ and $\mathcal{L}_{\mathrm{distill}}$, on both detection accuracy and scan sparsity. Table \ref{tab:ablation_loss} demonstrates that without these losses, both performance and sparsity suffer significantly. Introducing the distillation loss, $\mathcal{L}_{\mathrm{distill}}$, alone yields substantial gains as it guides the predicted queries to mimic full‐scan queries. The CVaR loss, $\mathcal{L}_{\mathrm{CVaR}}$, further enhances robustness by specifically forcing the model to improve its predictions on small and rare objects. Combining both losses yields the best detection performance with a slight decrease in sparsity. This is an expected trade-off, as $\mathcal{L}_{\mathrm{CVaR}}$ directs the model to scan more diverse areas.

%Specifically, $\mathcal{L}_{\mathrm{distill}}$ guides predicted queries to align closely with full-scan queries, ensuring informative regions are identified effectively. Meanwhile, $\mathcal{L}_{\mathrm{CVaR}}$ emphasizes accurate prediction for smaller and rare objects. Introducing $\mathcal{L}_{\mathrm{distill}}$ alone notably improves both performance and sparsity. However, incorporating both losses achieves optimal performance, albeit with a slight decrease in sparsity. This minor sparsity reduction occurs because $\mathcal{L}_{\mathrm{CVaR}}$ encourages thorough coverage of infrequent yet critical objects, thus enhancing overall accuracy at the expense of a modest increase in scan density. 
\begin{table}[!hbp]
  \centering
  \footnotesize
  \renewcommand{\arraystretch}{0.9}
   \vspace{-2.5mm}
  \caption{Impact of loss terms on nuScenes validation set}
  \vspace{-3mm}
  \label{tab:ablation_loss}
  \resizebox{0.75\columnwidth}{!}{%
    \begin{tabular}{@{}c|c|c|c|c@{}}
      \toprule
      $\bm{\mathcal{L}}_{\mathrm{CVaR}}$
       & $\bm{\mathcal{L}}_{\mathrm{distill}}$
        & \textbf{mAP\% $\uparrow$}
        & \textbf{NDS\% $\uparrow$}
        & \textbf{Sparsity\% $\uparrow$}
        \\
      \midrule
                    &    &  66.5  &  69.9         &   65.4 \\
      \checkmark     &    &  67.0  & 70.3 &  65.8 \\
       &   \checkmark &  67.3  &  70.4         & \textbf{66.3} \\
      \checkmark      & \checkmark      & \textbf{67.5}     & \textbf{70.6}            & 66.0      \\
      \bottomrule
    \end{tabular}%
  }
  \vspace{-5mm}
\end{table}
%\vspace{-2mm}
\paragraph{Impact of Differentiable Voxelization:} Table \ref{tab:ablation_voxxel} evaluates the effect of our proposed differentiable voxelization that allows the gradient propagating through the end-to-end datapath. %Traditional voxelization methods block gradient flow between the detector and the Mask Generator, potentially overlooking regions important for detection (i.e. areas surrounding object boundaries). In contrast, the differentiable voxelization enables gradient flow, guiding the model to select informative regions beyond object bounding boxes. As shown, 
It confirms that, without differentiable voxelization, the detection performance drops significantly. Incorporating the proposed approach notably improves accuracy, though sparsity slightly decreases. This minor reduction in sparsity is expected, as the model now identifies additional ROI areas that are critical to the task performance.
\begin{table}[htbp]
  \centering
  \footnotesize
  \renewcommand{\arraystretch}{0.85}
  \vspace{-2.5mm}
  \caption{Impact of differentiable voxelization on nuScenes}
  \vspace{-3.5mm}
  \label{tab:ablation_voxxel}
  \resizebox{0.7\columnwidth}{!}{%
    \begin{tabular}{@{}c|c|c|c@{}}
      \toprule
      \textbf{Diff. voxel}
      & \textbf{mAP\% $\uparrow$}
        & \textbf{NDS\% $\uparrow$}
        & \textbf{Sparsity\% $\uparrow$}
        \\
      \midrule
                  &  66.7  &  70.0         &   68.1 \\
         \checkmark   & \textbf{67.5}     & \textbf{70.6}            & 66.0 \\
      \bottomrule
    \end{tabular}%
  }
  \vspace{-5mm}
\end{table}
\section{Conclusion}
\label{sec:Conclusion}
We have introduced an adaptive LiDAR scanning framework that enhances sensor fusion systems' efficiency by leveraging temporal history to predict and selectively scan salient regions of a scene. Our two-stage approach uses a history-aware Query Predictor to predict object locations and a differentiable Mask Generator to create an efficient scanning policy. Experiments on nuScenes and Lyft datasets demonstrate that this method can reduce LiDAR scanning by over $65\%$, achieving competitive performance or outperforming dense-scanning baselines, all while reducing computational overhead. This work represents a significant step towards more intelligent, efficient, and sustainable perception systems for autonomous vehicles.
\section{Acknowledgments}
This work was supported in part by COGNISENSE, one of seven
centers in JUMP 2.0, a Semiconductor Research Corporation (SRC) program sponsored by DARPA.

\bibliographystyle{aaai2026}
\bibliography{aaai2026}
\appendix
\newpage
\begin{table*}[t]
  \centering
  \footnotesize
  \renewcommand{\arraystretch}{1} % slightly taller rows
  \caption{Performance comparison for next-frame prediction on nuScenes validation set. C.V., Ped., M.C., B.C., T.C., and B.R. represent
construction vehicle, pedestrian, motorcycle, bicycle, traffic cone, and barrier, respectively. Scan sparsity = percentage of empty pixels. }
  \label{appendix:nusc-nextframe}
  \vspace{-2mm}
  \begin{tabular}{@{}l|c|c|c|c c c c c c c c c c@{}}
    \toprule
    \textbf{Model}  &\textbf{mAP\% $\uparrow$\,} & \textbf{NDS\% $\uparrow$\,} & \textbf{\shortstack{Scan\\\\Sparsity\% $\uparrow$}} & \textbf{Car} & \textbf{Truck} & \textbf{Bus} & \textbf{Trailer} & \textbf{C.V.} & \textbf{Ped.} & \textbf{M.C.} & \textbf{B.C.} & \textbf{T.C.} & \textbf{B.R.}\\
    \midrule

    CMT (800×320) & 67.9          & 70.7   & 0 & 87.7 & 64.4 & 76.0 & 45.3 &  31.4 & 86.2 & 76.4 & 65.4 & 75.1 & 71.5 \\
    \rowcolor{gray!15}
    +Adaptive Scan & 67.5 & 70.6   & \textbf{66.0} & 87.5  & \textbf{64.5} & 74.9 & 43.7 & 29.6 & 85.5 & 75.9  & 65.2 & 75.0 & \textbf{72.9}\\
    CMT (1600×640) & 70.3 & 72.9 & 0 & 88.8 & 65.5 & 78.3 & 46.4 & 33.9 & 87.8 & 79.6 & 70.2 & 79.4 & 73.1\\
    \rowcolor{gray!15}
    +Adaptive Scan & \textbf{71.0} & \textbf{73.1} & \textbf{66.8} & 88.3 & \textbf{67.9} & \textbf{79.9} & \textbf{48.7} & 33.3 & 87.1 & 78.5 & \textbf{71.2} &  78.7 & \textbf{76.4}\\
    \bottomrule
  \end{tabular}
  \vspace{-2mm}
\end{table*}

\begin{table*}[t]
  \centering
  \footnotesize
  \renewcommand{\arraystretch}{1} % slightly taller rows
  \caption{Temporal performance comparison for the entire sequence prediction on nuScenes validation set.}
  \label{appendix:nusc-temporal}
  \vspace{-2mm}
  \begin{tabular}{@{}l|c|c|c|c c c c c c c c c c@{}}
    \toprule
    \textbf{Model} &\textbf{mAP\% $\uparrow$\,} & \textbf{NDS\% $\uparrow$\,} & \textbf{\shortstack{Scan\\\\Sparsity\% $\uparrow$}} & \textbf{Car} & \textbf{Truck} & \textbf{Bus} & \textbf{Trailer} & \textbf{C.V.} & \textbf{Ped.} & \textbf{M.C.} & \textbf{B.C.} & \textbf{T.C.} & \textbf{B.R.}\\
    \midrule

    CMT( 800×320) & 67.9          & 70.7   & 0 & 87.7 & 64.4 & 76.0 & 45.3 &  31.4 & 86.2 & 76.4 & 65.4 & 75.1 & 71.5 \\
    \rowcolor{gray!15}
    +Adaptive Scan   & 67.5 & 70.6   & \textbf{65.3} & 87.4  & \textbf{64.6} & 74.9 & 44.0 & 29.6 & 85.4 & 75.5  & \textbf{65.9} & \textbf{75.2} & \textbf{72.5}\\
    CMT (1600×640) & 70.3 & 72.9 & 0 & 88.8 & 65.5 & 78.3 & 46.4 & 33.9 & 87.8 & 79.6 & 70.2 & 79.4 & 73.1\\
    \rowcolor{gray!15}
    +Adaptive Scan & \textbf{71.0} & \textbf{73.1} & \textbf{66.4} & 88.4 & \textbf{68.0} & \textbf{79.7} & \textbf{48.6} & 33.4 & 87.2 & 78.8 & \textbf{71.3} &  78.7 & \textbf{76.2}\\
    \bottomrule
  \end{tabular}
  \vspace{-3mm}
\end{table*}

\begin{figure*}[!htbp]
 \centering
 \includegraphics[width=2.1\columnwidth]{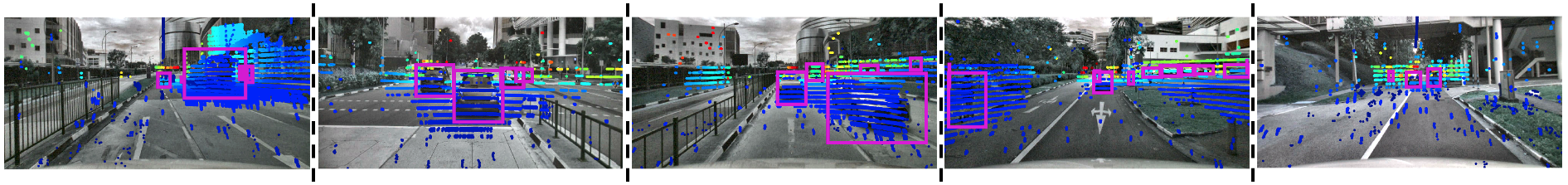}
 %\vspace{-0.75\baselineskip}
 \centering
 \vspace{-2mm}
 \caption{Visualizations of our adaptive LiDAR scanning method on different frames from the nuScenes validation set. These results were generated using the CMT model with an image resolution of 1600×640. Each sub-figure displays the selected LiDAR points, which are densely scanned within regions of interest and sparsely sampled elsewhere. The color of each point encodes its distance from the ego-vehicle.}
 \vspace{-0.6\baselineskip}
 \label{appendix:nusc}
\end{figure*}
\begin{figure*}[!htbp]
 \centering
 \includegraphics[width=2.1\columnwidth]{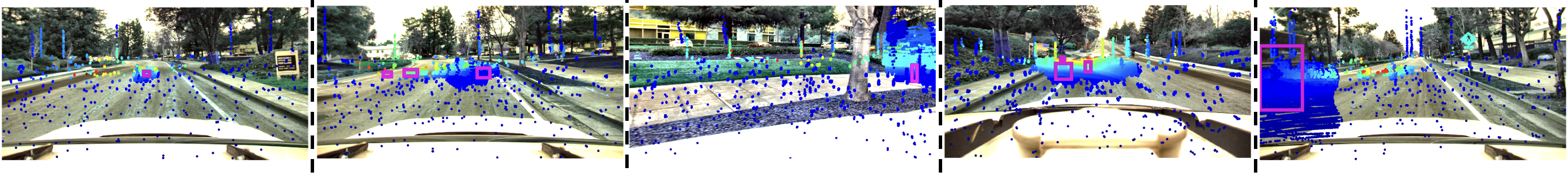}
 %\vspace{-0.75\baselineskip}
 \centering
 \vspace{-3mm}
 \caption{Visualizations of our adaptive LiDAR scanning method on different frames from the Lyft validation set.}
 \vspace{-1\baselineskip}
 \label{appendix:lyft}
\end{figure*}
\section{Appendix}
This supplementary section provides a detailed description of our model implementation, complementing the information presented in the paper.

\subsection{Implementation Details}

\paragraph{Model Optimization and Training Setting:} Our adaptive scanning module is integrated into the Cross-Modal Transformer (CMT)\cite{yan2023cross} framework. The transformer detection head is configured with $6$ decoder layers and a fixed number of $900$ object queries. For our history-aware prediction module, we use a temporal buffer of $T=4$ frames. The image stream is processed by a ResNet-50 \cite{he2016deep} backbone for low-resolution ($800\times320$) experiments and a VoVNet-99 \cite{lee2020centermask} for high-resolution ($1600\times640$) experiments. The LiDAR stream is processed through a voxel-based pipeline. The point cloud is first converted into a voxel grid using a voxel feature encoder \cite{zhou2018voxelnet} that retains the raw point features. These voxel features are then processed by a 3D sparse convolutional middle encoder followed by a 2D convolutional backbone based on the SECOND \cite{yan2018second} architecture with a Feature Pyramid Network (FPN) neck. All models are trained using the AdamW optimizer with a weight decay of $0.01$ and a cyclic learning rate schedule. We use mixed-precision training with dynamic loss scaling and clip the gradient norm at a maximum value of $50$. All experiments are conducted on $8$ NVIDIA A40 GPUs with a batch size of $2$ and $9$ per GPU for low and high image resolution, respectively.
\paragraph{Data Augmentation: }
During training, we apply several data augmentation techniques to improve model robustness. Both the LiDAR point clouds and the multi-view images undergo random horizontal flipping. However, global rotation and scaling augmentations are not used in our inference experiments. For the image stream, we perform a series of transformations that include resizing the original images and taking a rectangular crop, which is then scaled to the final target input resolution (e.g., $800\times320$ or $1600\times640$). Following standard practice \cite{yan2023cross}, each image is then normalized using a mean of $[103.530,116.280,123.675]$ and a standard deviation of $[57.375,57.120,58.395]$ for both datasets. Furthermore, to enhance the robustness of the fusion model, we employ the masked-modal training strategy from CMT. In this process, we randomly `mask out' an entire sensor modality (either camera or LiDAR) for a given training sample with a fixed probability, forcing the model to learn strong uni-modal representations and preventing it from becoming overly reliant on a single sensor stream.
\paragraph{nuScenes Low-Resolution Configuration: } For this setup, we use an input image resolution of $800\times320$. The point cloud detection range is set to $[-54\,\mathrm{m},\,54\,\mathrm{m}]$ for the X/Y axes and $[-5\,\mathrm{m},\,3\,\mathrm{m}]$ for the Z axis with a voxel size of $[0.1\,\mathrm{m},0.1\,\mathrm{m},0.2\,\mathrm{m}]$ meters. The model is trained for $21$ epochs with a base learning rate of $1\times10^{-4}$. We apply learning rate multipliers of $0.01$ to the image backbone and $0.1$ to the image neck for stable training.
\paragraph{nuScenes High-Resolution Configuration: } For the high-resolution experiments, the input image resolution is $1600\times640$. To accommodate the denser features, the LiDAR voxel size is reduced to $[0.075\,\mathrm{m},0.075\,\mathrm{m},0.2\,\mathrm{m}]$, while the point cloud range remains the same. This configuration is also trained for $21$ epochs, but with a base learning rate of $5\times10^{-5}$. The learning rate multipliers for the image backbone and neck remain the same as in the low-resolution setting.
\paragraph{Lyft Level 5 Configuration: } The model is trained on Lyft's $9$ object classes for $21$ epochs. We use a larger point cloud range of $[-100\,\mathrm{m},\,100\,\mathrm{m}]$ for the X/Y axes and $[-5\,\mathrm{m},\,3\,\mathrm{m}]$ for the Z axis, with a voxel size of $[0.125\,\mathrm{m},0.125\,\mathrm{m},0.2\,\mathrm{m}]$. The image backbone is VoVNet-99 with an input resolution of $1600\times640$. The base learning rate is $5\times10^{-5}$. For this dataset, we found it beneficial to use different learning rate multipliers of $0.001$ for the image backbone, and $0.01$ for the image neck.
\paragraph{Mask Post-Processing Pipeline for ROIs Selection:} The raw probability map produced by our mask generator can be noisy, occasionally containing spurious activations or holes within larger regions of interest. To produce a spatially coherent and robust scanning policy, this raw output is therefore refined through a post-processing pipeline. Our post-processing adopts hysteresis thresholding \cite{sornam2016hysteresis} to obtain de-noised and contiguous ROIs that are scanned densely. Then, for all areas outside these ROIs, it applies a stochastic Bernoulli selection with a fixed probability to create a sparse scanning pattern.

The first step employs hysteresis thresholding \cite{sornam2016hysteresis} to clean the initial prediction by leveraging both high and low confidence thresholds. This process begins by automatically determining a high confidence threshold on the probability map using Otsu's method \cite{barron2020generalization}. Pixels with probabilities above this value are identified as high-confidence `seeds'. A second, much lower threshold is then used to define a broader `support' region of all potentially relevant pixels. The final mask is generated by performing a morphological reconstruction, where the initial seed regions are grown to include any connected pixels within the support region. This procedure effectively fills gaps between high-confidence areas and eliminates isolated noise. To complete the cleanup, small, disconnected objects and any remaining holes within the grown mask are removed using additional morphological operations.

For all areas outside of these ROIs, we apply a sparse scanning pattern to maintain situational awareness without significant energy cost. This sparse pattern is generated via quantized Bernoulli sampling. The original low-confidence probability values in these non-ROI regions are first quantized by snapping them to a predefined set of sampling rates (e.g., $[0.0625,0.125,0.25,0.5,1.0]$). A final binary decision is then made for each non-ROI pixel by performing Bernoulli sampling according to the quantized probability. The final scan pattern is a composite of the dense ROI mask and this sparse non-ROI mask, ensuring that sensor energy is concentrated on the most important parts of the scene while still monitoring the entire environment.
\subsection{Performance Breakdown for Each Category on nuScenes }
Tables \ref{appendix:nusc-nextframe} and \ref{appendix:nusc-temporal} present the per-category performance of our adaptive LiDAR scanning on the nuScenes validation split \cite{caesar2020nuscenes}. When computing LiDAR sparsity, the first four frames ($T=4$) used to initialize the temporal buffer are excluded. For next-frame prediction (Table \ref{appendix:nusc-nextframe}), our approach achieves $66.8\%$ LiDAR sparsity in the high-resolution setting while outperforming the CMT baseline in both mAP and NDS. These gains are consistent across most classes, with the largest improvements observed for `barriers' and `buses'. In the low-resolution setting, our approach achieves $66.0\%$ LiDAR sparsity while preserving competitive detection accuracy with the baseline. A similar trend is observed for the entire sequence prediction (Table \ref{appendix:nusc-temporal}), where our adaptive scanning method demonstrates robust accuracy at high levels of sparsity.

\subsection{Visualization of Adaptive LiDAR Scanning}
We visualize adaptive scanning results on selected frames from the nuScenes validation set and the Lyft validation set. Figure \ref{appendix:nusc} displays the regions chosen by our adaptive scanning module when integrated with CMT at an image resolution of $1600 \times 640$. Figure \ref{appendix:lyft} presents the same results on the Lyft validation set. In each region of interest, the model densely samples LiDAR points and randomly samples points outside these regions. We also observe that the module selects points outside the annotated boxes, illustrating how differentiable voxelization enables the policy network to leverage surrounding contextual information that can enhance detection accuracy.
\subsection{Visualization of 3D Object Detection}
\begin{figure*}[!htbp]
 \centering
 \includegraphics[width=2.1\columnwidth]{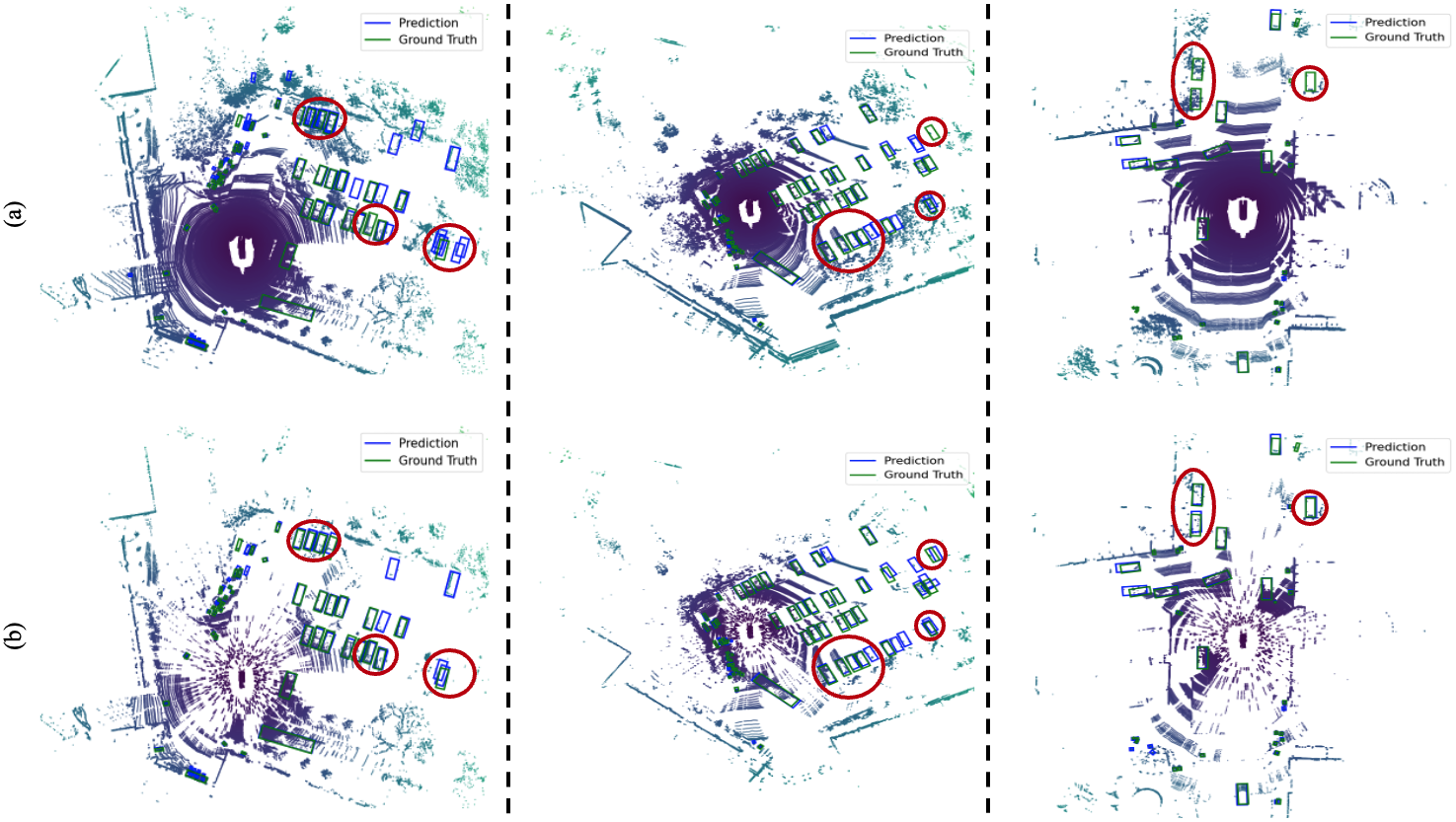}
 %\vspace{-0.75\baselineskip}
 \centering
 \vspace{-2mm}
 \caption{Comparison of 3D object detection results using dense LiDAR scanning (a) and proposed adaptive sparse scanning (b) on the nuScenes validation set.  Blue and green boxes are the prediction and ground truth boxes. It can be seen that the adaptive scanning module achieves detection results comparable to full scanning and even improves performance in certain challenging cases highlighted by red circles.}
 \vspace{-0.6\baselineskip}
 \label{appendix:BEV_nusc}
\end{figure*}

\begin{figure*}[!htbp]
 \centering
 \includegraphics[width=2.1\columnwidth]{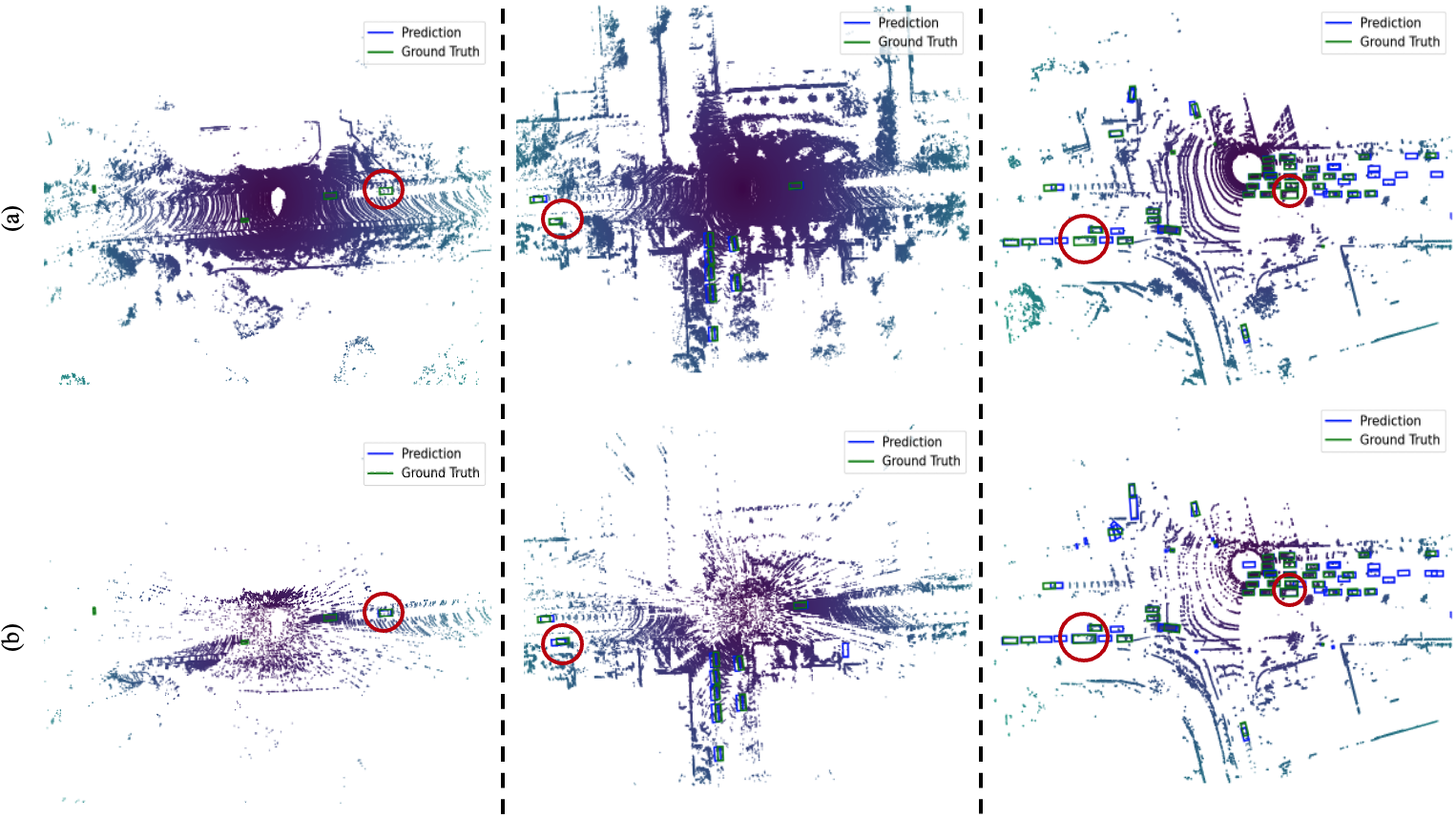}
 %\vspace{-0.75\baselineskip}
 \centering
 \vspace{-2mm}
 \caption{Comparison of 3D object detection results using dense LiDAR scanning (a) and proposed adaptive sparse scanning (b) on the Lyft validation set. Our proposed method can outperform the dense LiDAR scanning in some of the hard object scenarios highlighted by red circles.}
 \vspace{-0.6\baselineskip}
 \label{appendix:BEV_Lyft}
\end{figure*}

We present a qualitative comparison of 3D object detection results produced by CMT using both dense and adaptive sparse scanning on selected frames from the nuScenes validation set, as shown in Figure \ref{appendix:BEV_nusc}. Our adaptive method achieves detection performance comparable to that of the full scan, and in some cases, successfully detects challenging objects missed by the conventional dense approach (highlighted in red). Similarly, Figure \ref{appendix:BEV_Lyft} demonstrates that our method performs on par with dense LiDAR scanning—and occasionally surpasses it (highlighted in red)—in several challenging scenarios on the Lyft dataset.

\end{document}